\documentclass[3p]{elsarticle}             
\usepackage{graphicx}    
\usepackage{xcolor}
\usepackage{amsmath,amssymb,amsfonts}
\usepackage{algorithmic}
\usepackage{textcomp}

\usepackage[ruled]{algorithm2e}

\begin{document}

\begin{frontmatter}

\title{Benchmarking federated strategies in Peer-to-Peer Federated learning for biomedical data}

\author[jl1,jl2]{Jose L. Salmeron}\ead{joseluis.salmeron@cunef.edu}
\author[irina]{Irina Ar\'evalo}\ead{iarebar@alu.upo.es}
\author[celma]{Antonio Ruiz-Celma}\ead{aruiz@unex.es}

\address[jl1]{CUNEF Universidad (Madrid, Spain)}
\address[jl2]{Universidad Aut\'onoma de Chile (Chile)}
\address[irina]{Universidad Pablo de Olavide (Seville, Spain)}  

\address[celma]{Universidad de Extremadura (Badajoz, Spain)}

\begin{keyword}                          
Federated learning, privacy-preserving machine learning.          
\end{keyword}

\begin{abstract}                         
The increasing requirements for data protection and privacy has attracted a huge research interest on distributed artificial intelligence and specifically on federated learning, an emerging machine learning approach that allows the construction of a model between several participants who hold their own private data. In the initial proposal of federated learning the architecture was centralised and the aggregation was done with federated averaging, meaning that a central server will orchestrate the federation using the most straightforward averaging strategy. This research is focused on testing different federated strategies in a peer-to-peer environment. The authors propose various aggregation strategies for federated learning, including weighted averaging aggregation, using different factors and strategies based on participant contribution. The strategies are tested with varying data sizes to identify the most robust ones. This research tests the strategies with several biomedical datasets and the results of the experiments show that the accuracy-based weighted average outperforms the classical federated averaging method.
\end{abstract}

\end{frontmatter}

\section{Introduction}
\footnote{Preprint - Published in Heliyon 9 (2023) e16925}
Artificial intelligence applications in healthcare are increasing every day. These applications have the ability to advance the healthcare industry by, for instance, supporting clinical decision making, risk prediction, developing early warning systems for patients, increasing the accuracy and timeliness of diagnosis, improving patient–physician interaction, and optimizing operations and resource allocation \cite{rahimi.2021}.

Federated learning is a new approach for distributed artificial intelligence that aims to have several agents train a deep learning model in a collaborative and secure way, without sharing any private data. This training is done the following way: a central server defines a deep learning model and sends it to the agents, who train the model in their private data. Then, they send the parameters of the model (weights or gradients) back to the server, who aggregates these data in order to find a global federated model, which in turn is delivered back to the agents to be retrained in their data. This process is iterated until convergence. 

In the initial definition of the federated learning approach, the aggregation step is done by averaging the model parameters. Nevertheless, other aggregation methods may be of more interest since they can improve the performance of the model by giving more weight to different agents depending on their size or the performance of the local models in their data. 

The main contributions of this research are two-fold: 
\begin{enumerate}
    \item Several aggregation strategies are proposed, such as weighted averaging aggregation using the dataset size, weighted average using the normalized inverse of the local test accuracy, weighted averaging aggregation using the dataset size and accuracy, weighted average using the contribution of the participant, and weighted sum using the inverse contribution of the participant. Federated averaging is included for comparison. 
\item The strategies are tested with different data sizes on each participant. This allows analyzing the strategies under different circumstances and identifying those that are more robust.
\end{enumerate}

The rest of this paper is organized as follows. We discuss the theoretical background of federated learning in section \ref{background}. The different federated strategies are described in section \ref{aggregations}. The methodological proposal can be found in section \ref{meth}, while section \ref{experiments} shows the results of the experimental approach, serving as a benchmark. Finally, the authors draw a conclusion in section \ref{conclusions}.

\section{Fundamentals} \label{background}
\subsection{Related work}

Federated learning is an emerging approach for distributed artificial intelligence in which the different data owners (or participants) train collaboratively a machine learning model \cite{liu.2020, salmeron.2020}. The model is updated (trained) in the own private data of each participant and then the trained model is sent for aggregation to a central server or one of the participants. It was first proposed by McMahan et al. \cite{mcmahan.2016} and further developed in Konecny et al. \cite{konecn.2016} and McMahan and Ramage \cite{mcmahan_ramage.2017}. 
The main advantage of federated learning is the training of a model with the private data of each participant keeping the security and compliance requirements while improving their models \cite{ahmed.2021}. 
It also allows the use of more accurate models with low latency, ensuring privacy and less power consumption \cite{yang.2019}.

Numerous surveys and literature reviews have extensively examined the body of work documented in academic literature pertaining to architectures, approaches, utilization, and applications of federated learning. In the healthcare domain, Hoyos et al. \cite{hoyos.2023} present federated learning approaches (horizontal, vertical and transfer learning) for FCMs for the prediction of mortality and the prescription of treatment of severe dengue. Antunes et al. \cite{antunes.2022} outline a broad architecture for federated learning applied to healthcare data, drawing upon key insights derived from the literature review. Li et al. \cite{li.2023} analyse recent literature on the utilization of federated learning, outlining various federated learning architectures and classification models. Nguyen et al. \cite{nguyen.2022} provide a comprehensive and up-to-date review of the latest advancements in federated learning within crucial healthcare domains, encompassing health data management, remote health monitoring, medical imaging, and COVID-19 detection. Xu et al. \cite{xu.2021} provide an overview of the common solutions to address statistical challenges, system challenges, and privacy issues in federated learning. They also highlight the potential implications and opportunities that federated learning holds for the healthcare sector. Furthermore, numerous studies have sought to optimize federated learning and broaden its practical applications, with research efforts spanning areas such as computation fusion \cite{zhao.2022}, data transmission \cite{sattler.2020, su.2021}, as well as privacy and security-related concerns \cite{hou.2022}.

\subsection{Architecture}

\begin{figure*}[ht]
\centering
\includegraphics[width=6in]{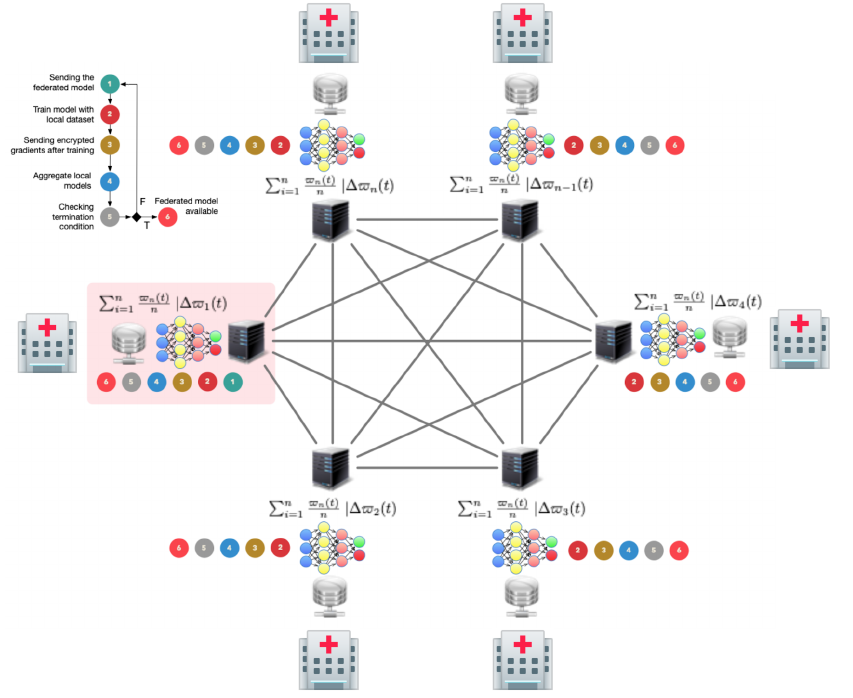}
\caption{Fully connected peer-to-peer federated learning architecture} 
\label{figureFL}
\end{figure*}

Fully decentralised learning aims to replace server-based communication with peer-to-peer communication among individual clients as its core concept. Each round in fully decentralized algorithms involves a client performing a local update and sharing information with their neighbors in the graph. In this research, clients share the model with all participants, making it a fully connected peer-to-peer architecture (see Figure \ref{figureFL}). 

Peer-to-peer architectures can have a significant impact on federated learning. With peer-to-peer communication, clients can collaborate and share locally trained models directly with each other, which can result in faster and more efficient learning compared to a centralized architecture. Peer-to-peer architectures can also provide better privacy and security as the clients can keep their data locally and only share the necessary information with their peers. 

Recently, a fully decentralised solution where participants collaborate asynchronously and communicate in a peer-to-peer fashion, without any central server to orchestrate a global state of the system or even to coordinate the protocol, is proposed in \cite{bellet.2018}. In this scenario, peer-to-peer architectures can also have some challenges. For example, it can be more difficult to coordinate and manage the communication between clients, and it may require additional mechanisms to ensure the consistency of the model across all clients. To overcome this challenge, this paper focuses on a fully connected peer-to-peer architecture. Moreover, this kind of systems may not be suitable for large-scale federated learning scenarios due to the high communication overhead between clients. However, the goal of this research is to present a fault-tolerant architecture in case of failure of a central server and/or in multiple participants.

In this case, the group of autonomous peers run iteratively multiple training rounds to update the federated model \cite{bellet.2018, wink.2021}. Indeed, peer-to-peer algorithms include scalability-by-design to large sets of devices thanks to the locality of their updates \cite{kermarrec.2015}. In addition, a decentralised peer-to-peer architecture intrinsically provides an additional some security guarantee as it becomes much more tough for any third party to get the full state information of the system \cite{wink.2021}.

A peer-to-peer federated learning process needs a minimum of two participants. In this case, there is not a central server managing the federated model and the communications with the participants. In a peer-to-peer architecture, each participant receives the trained models of the remainder participants and then averages the participants' models to obtain a federated model trained with the private data. The participants own the data and train the partial models. This process can be iterated as many times as needed. The process is shown in Figure \ref{figureFL} and is as follows:

\begin{enumerate}
\item A participant initiates the federated learning process by sending an initial model to all the other participants. If this is the initial iteration, the federated model is dispatched by the participant triggering the process.
\item Each participant trains the received model using their own private data.
\item Each participant sends the parameters of the model in a private way (usually encrypting the data to be sent) to the remaining participants.
\item Every participant aggregates the partial models with the updated parameters and builds the federated model according to a federation strategy.
\item Every participant checks a termination condition in either accuracy of the model in a test dataset or number of iterations. If it is accomplished, the federated learning process ends, otherwise the process iterate again from step 1.
\end{enumerate}

The target of the federated learning process is to minimize the total loss for all participants, computed as in Equation \ref{eq:loss},

\begin{equation}\label{eq:loss}
     \mathcal{L}^* = \frac{1}{n} \sum_{i=1}^n \mathcal{L}_i(\mathcal{D}_i,\Phi)
\end{equation}
\noindent where $\mathcal{L}^*$ is the loss function for the federated model, $\mathcal{L}_i$ is the loss function for each participant in the federation, $\mathcal{D}_i$ is the dataset of the participant $i$ and $\Phi$ is the federated model parameters, $n$ is the number of participants.

The federated learning process trains a model between different participants without the sharing of private data. Nevertheless, there are other possible risks, like model poisoning, potential attacks to reconstruct the model or the training data from the parameters that the participants send to the central server, or the use of attack models \cite{pmlr-v108-bagdasaryan20a, Wang19}. As a consequence, there have been several advances in the use of privacy-preserving methods such as Differential Privacy or Homomorphic Encryption in federated learning, see \cite{abadiDP16, Acar17, Kaissis2020SecurePA, Hu20, Cheng20}. 

%
%
%

\subsection{Federation strategies}\label{aggregations}

As mentioned before, the aggregation method for federated learning is an important parameter of the process. The original definition contemplated the arithmetic mean of the parameters of the model to obtain the federated model. 

In this research, the authors propose a series of different aggregations and compare them in our experimental section (see section \ref{experiments}). Assuming that the parameters of the model at iteration $j$ is as shown in Equation \ref{eq:pars},
\begin{equation}\label{eq:pars}
\Phi_j = [\Phi_{j1}, \Phi_{j2}, \cdots, \Phi_{jn}]
\end{equation}
\noindent where $n$ is the number of participants, $\mathcal{D}_i$ is the dataset of the participant $i$, and $\Phi_j'$ is the parameters (weights or gradients) of the federated model, the functions of the parameters that we will discuss are the following: 
\begin{itemize}
    \item Average of the parameters (weights or gradients):
    \begin{equation}\label{average}
    \Phi_j' = \frac{1}{n}\sum_{i=1}^n \Phi_{ji}
    \end{equation}
    where every participant contributes the same to the global model.
    \item Weighted averaging aggregation using the normalized size of each participants' dataset:
    \begin{equation}\label{size}
    \Phi_j' = \sum_{i=1}^n \frac{|\mathcal{D}_i|}{\sum_{k = 1}^n |\mathcal{D}_k|}\cdot\Phi_{ji}    
    \end{equation}
    \noindent where every participant contributes to the global model proportionally to the size of their data, and agents with less information will affect less to the final model.
    \item Weighted average using the normalized inverse accuracy of the model in a test set of each participants: 
    \begin{equation}\label{accuracy}
    \Phi_j' = \sum_{i=1}^n \frac{1/\textrm{acc}_{ji}}{\sum_{k = 1}^n 1/\textrm{acc}_{jk}}\cdot\Phi_{ji}
    \end{equation}
    \noindent where the individual models add to the global model inversely to their performance metric, trying to give more weight to the less accurate models in order to improve their metric in their datasets. 
    \item Weighted average using the accuracy and the size of the dataset: 
    \begin{equation}\label{accsize}
    \Phi_j' = \sum_{i=1}^n \frac{\textrm{acc}_{ji}|\mathcal{D}_i|}{\sum_{k = 1}^n |\mathcal{D}_k|}\cdot\Phi_{ji}
    \end{equation}
    \noindent where the contribution of each participant's model depends on both the accuracy of the model and the size of the dataset.
    \item Weighted average using the contribution ($\mathcal{C}$) of the participant, that is, the normalized inverse of the absolute difference between the loss of the participant's model and the loss of the global model when applied to the participants' data as shown in Equation \ref{eq:contrib}
    \begin{equation}\label{eq:contrib}
    \mathcal{C}_{ji}=|\mathcal{L}_j^*(\mathcal{D}_i, \Phi)-\mathcal{L}_j(\mathcal{D}_i, \Phi)|
    \end{equation}
    and 
    \begin{equation}\label{contribution}
    \Phi'_j = \sum_{i=1}^n \frac{\mathcal{C}_{j,i}}{\sum_{k=1}^n \mathcal{C}_{jk}}\cdot\Phi_{ji}
    \end{equation}
    \item Weighted sum using the inverse contribution of the participant: 
    \begin{equation}\label{invcont}
    \Phi'_j = \sum_{i=1}^n \frac{1/\mathcal{C}_{ji}}{\sum_{k = 1}^n 1/\mathcal{C}_{jk}}\cdot\Phi_{ji}  
    \end{equation}
\end{itemize}

\section{Methodological proposal}\label{meth}

The federated learning proposal starts with a central server sending an untrained model to the participants. As a first step, each hospital trains this model with their training data, evaluates it in their test data, and sends the parameters of the trained model back to the server. After receiving all the parameters from all the participants, in this proposal the server aggregates the parameters using one of the aggregation methods described in Section \ref{aggregations} to obtain the global method. This process is iterated until convergence.
In this use case the authors have proposed a peer-to-peer architecture, and have not included an additional encryption layer, but it is possible and desirable to do so in real-life applications in healthcare.

\begin{figure}[ht]
\centering
\includegraphics[width=2.5in]{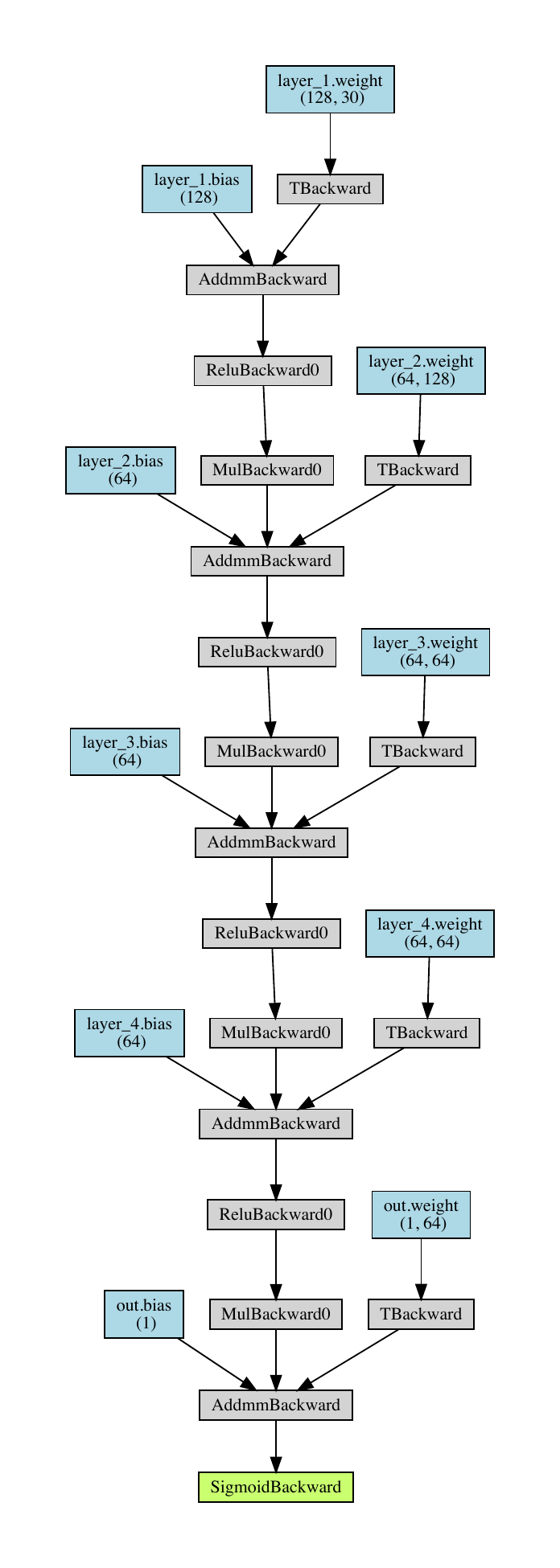}
\caption{Deep Neural Network topology for the experiments} 
\label{NN}
\end{figure}

In the following experiments, the initial model will be a dense neural network made of five layers followed by a non-linear ReLU function 
and a dropout layer for regularization (Figure \ref{NN}). For training, the loss function will be computed using the Binary Cross Entropy. 

As a setup for the experiments, the authors assume that several hospitals with their own private data wish to train a deep learning model for diagnosis of a disease, but the size of their data is not large enough for training an accurate model. The data of all the hospitals should not be combined due to data regulations given their special sensitivity. Therefore, for each one of the experiments the corresponding dataset will be randomly split in five different subsets, to simulate five hospitals. In a first test all of them will have the same amount of data, obtained by splitting evenly the dataset. The other three test will be random splits among the participants, where in the last two we have forced that there would be several participants with a very small number of samples (less than 10\%). The distribution of the variables, including the percentage of positive cases for the target, will also vary from one hospital to another.

In every case, each participant's data will be split into a train and a test dataset for training and evaluation. 

The metric we will use to compare the performance of the different models and methods is the accuracy on the test set. Given a classification model, its performance can be summarized with a confusion matrix, a table that shows the number of real positive and negative samples in our dataset versus the number of positive and negative values that our classifier predicts. 


The accuracy of a model is the ratio of number of correct predictions to the total number of input samples,
that is, the sum of the main diagonal of the confusion matrix divided by the sum of all values in the table, which is the amount of times the classifier got the prediction right. This metric varies between 0 and 1, where 0 is the worst case scenario, and 1 is associated to a perfect classifier. Moreover, a random binary classifier such as a coin toss has an accuracy of 0.5, and therefore this is a first baseline for any balanced binary classifier, understanding that an accuracy close to 0.5 is as bad as a random guess.

The results of the experiments will be shown as follows: for each experiment, each line will represent one of the five participant hospitals. The columns represent the number of the partition, its size, and the percentage of positives in the partition, and the accuracy of a model evaluated in each participant's test set. The first one is the local model in every participant without any federation, and the following are the results for each aggregation method described in Section \ref{aggregations}: Federated Average (Equation \ref{average}) and weighted sums with size (Eq. \ref{size}), inverse accuracy (Eq. \ref{accuracy}), size and accuracy (Eq. \ref{accsize}), contribution (Eq. \ref{contribution}), and inverse contribution (Eq. \ref{invcont}). The average accuracy for all participants in each experiment will also be included and used for comparison of the aggregation methods. 

Table \ref{column.names} summarizes how the results of the experiments will be shown, and what federation strategies are being evaluated.

\begin{table*}[!ht]
\small
\centering
\caption{Column names and federation strategies}\label{column.names}
\begin{tabular}{| l | l | c |}
\hline
Column name & Definition & Federation strategy \\
\hline
\hline
Part & Partition number & \\
\hline
Size & Percentage of the  & \\
 & total samples in this agent & \\
\hline
Pos (\%) & Percentage of positives in the partition &\\ 
\hline
Acc (local) & Accuracy in the non-federated case & \\
\hline
Acc Federated & Accuracy of the federated model & Equation \ref{average} \\
Averaging (Fed) & using fed averaging &  \\\hline
Acc. Size (Fed) & Accuracy of the model & Equation \ref{size} \\
& using size-based averaging & \\\hline
Acc. Accuracy (Fed) & Accuracy of the federated model & Equation \ref{accuracy}\\
& using accuracy-based averaging & \\\hline
Acc. Size \& Acc. (Fed) & Accuracy of the federated model & Equation \ref{accsize}\\
& using both accuracy and size in the averaging & \\\hline
Acc. Cont (Fed) & Accuracy of the federated model & Equation \ref{contribution}\\
& using the contribution of the participants & \\\hline
Acc. Inv. Cont. (Fed) & Accuracy of the federated model & Equation \ref{invcont}\\
& using the inverse contribution of the participants & \\\hline
\end{tabular}
\end{table*}

These results will serve as a benchmark for which averaging strategy improves the most the performance of the federated learning process, by comparing the accuracy in different datasets of the competing approaches. The benchmarking methodology is as follows:
\begin{enumerate}
    \item Define the problem: Firstly, define the problem that the AI model is intended to solve. In this research, we have selected four medical problems (breast cancer, chronic kidney disease, Parkinson's disease, and heart disease) to better validate our proposals.
    \item Identify the relevant data: We have selected well-known datasets for each of these medical problems to ensure reproducibility. 
    \item Preprocess and split the data: The datasets have already been preprocessed and are ready for use. The next step is to split the data into training and testing sets to evaluate the performance of the model. The training set will be used to train the model, while the testing set will be used to measure its accuracy on unseen data.
    \item Train the model: Train the Federated FCM model using different aggregation strategies.
    \item Evaluate the model: Evaluate the performance of the model using the test set using the accuracy metric.
    \item Compare the results: Conduct a comparison of the outcomes achieved by each aggregation strategy with the remaining ones to ascertain whether there are any significant performance differences among the strategies.
\end{enumerate}

\section{Experiments}\label{experiments}

\subsection{Experiment 1 - Breast Cancer}

The Breast Cancer Wisconsin Dataset contains descriptions of features of the nucleus of a breast mass, obtained from digitized images of the fine-needle aspirate, for 569 patients, where 212 are malignant and 357 are benign tumors. This dataset is publicly available \cite{dua.2019}. More details can be found in \cite{street.1993}, \cite{mangasarian.1995}.

The results of the federated learning process for this dataset are shown in Table \ref{results.breast}. As we can see, the federation improves the accuracy metrics in general, but there are differences between the aggregation methods used. In the first case of the even dataset all accuracies are improved, but in the first uneven split, the contribution aggregation of the difference of the losses does not induce an increase in the performance. In the second uneven split, more extreme than the previous one, with one participant with only 3\% of the data, the Federated Averaging does not improve the performance of the local models, and in the last one, with two participants with 6\% and 7\% of the data, the weighted aggregation using the size of each participant and the size and the accuracy show a lower accuracy than the first iteration of local models. 

Summing up, for this experiment the accuracy-based aggregation and the inverse contribution are the only aggregation methods that improve the performance of the local models after the federation process.

\begin{table*}[ht]
\centering
\scriptsize
\tiny
\caption{Experiment 1 - Breast Cancer}\label{results.breast}
\begin{tabular}{c | c | c|| c|| c|| c || c|| c|| c || c}
\hline
& & & \textbf{Acc} &         \textbf{Acc Federated} &        \textbf{Acc Size} &  \textbf{Acc Accuracy} & \textbf{Acc Size \& acc} &  \textbf{Acc Cont} & \textbf{Acc Inv Cont} \\
 \textbf{Part} & \textbf{Size} & \textbf{Pos (\%)}  & (\textbf{local}) & \textbf{Averaging} (\textbf{Fed}) & (\textbf{Fed}) &  (\textbf{Fed})  &  (\textbf{Fed}) & (\textbf{Fed}) &  (\textbf{Fed}) \\
\hline
1 & 20\% & 47\% & 0.5454 & 0.6363 &  0.6363  & 0.3636 & 0.7272 & 0.6363 & 0.7272\\
2 & 20\% & 41\% & 0.4166 & 0.5000 &  0.5833  & 0.7500 & 0.7500 & 0.6666 & 0.7500 \\
3 & 20\% & 31\% & 0.5454 & 0.7272 &  0.8181  & 0.5454 & 0.8181 & 0.6363 & 0.8181\\
4 & 20\% & 35\% & 0.5833 & 0.5833 &  0.6666  & 0.5833 & 0.7500 & 0.4166 & 0.8333\\
5 & 20\% & 32\% & 0.5833 & 0.6666 &  0.6666  & 0.7500 & 0.4166 & 0.5833 & 0.5833\\\hline
Avg & - & - & 0.5348 & \textbf{0.6227} & \textbf{0.6742} & \textbf{0.5984} & \textbf{0.6924} & \textbf{0.5879} & \textbf{0.7424}\\
\hline\hline
1 & 20\% & 12\% & 0.9090 & 0.9090 & 0.8000 & 0.8181 & 0.8888 & 0.8181 & 0.8181\\
2 & 22\% & 30\% & 0.6153 & 0.5384 & 0.7692 & 0.5333 & 0.5714 & 0.7058 & 0.5000\\
3 & 18\% & 47\% & 0.5000 & 0.6000 & 0.3636 & 0.4444 & 0.5454 & 0.4285 & 0.4000\\
4 & 17\% & 17\% & 0.7000 & 0.7000 & 0.9000 & 0.7500 & 0.7500 & 0.6363 & 0.8000\\
5 & 24\% & 39\% & 0.4285 & 0.5714 & 0.6666 & 0.6666 & 0.4285 & 0.3571 & 0.7142\\\hline
Avg & - & - & 0.6306 & \textbf{0.6637} & \textbf{0.6999} & \textbf{0.6425} & \textbf{0.6368} & 0.5892 & \textbf{0.6465}\\\hline\hline
1 & 49\% & 28\% & 0.5714 & 0.6428 & 0.6333  & 0.6896 & 0.7142 & 0.7096 & 0.6896\\
2 &  3\% & 40\% & 0.5000 & 0.0000 & 1.0000  & 0.6666 & 1.0000 & 0.5000 & 0.6666\\
3 & 15\% & 40\% & 0.8888 & 1.0000 & 0.5000  & 0.8333 & 0.6250 & 0.5000 & 0.8333\\
4 &  5\% & 63\% & 0.3333 & 0.3333 & 1.0000  & 0.6666 & 0.0000 & 0.6666 & 0.6666\\
5 & 29\% & 20\% & 0.4117 & 0.5882 & 0.7500  & 0.5789 & 0.7368 & 0.8666 & 0.5789\\\hline
Avg & - & - & 0.5410 & 0.5128 & \textbf{0.7766} & \textbf{0.6870} & \textbf{0.6152}  & \textbf{0.6486} & \textbf{0.6870} \\\hline\hline
1 & 48\% & 30\% & 0.6666 & 0.7333 & 0.7419  & 0.6969 & 0.6774 & 0.7000 & 0.8000  \\
2 &  7\% & 33\% & 0.6666 & 0.6666 & 0.7500  & 0.7500 & 0.5000 & 1.0000 & 0.6666 \\
3 &  6\% & 27\% & 0.6666 & 0.6666 & 0.8000  & 1.0000 & 0.5000 & 0.7500 & 0.3333 \\
4 & 16\% & 15\% & 1.0000 & 1.0000 & 0.3750  & 0.8000 & 0.8750 & 0.8750 & 1.0000 \\
5 & 23\% & 38\% & 0.4545 & 0.4545 & 0.5000  & 0.3636 & 0.5000 & 0.5454 & 0.7272 \\\hline
Avg & - & - & 0.6970 & \textbf{0.7042}  & 0.6334 & \textbf{0.7221}  & 0.6104 &  \textbf{0.7741} & \textbf{0.7055} \\\hline
\end{tabular}
\end{table*}

\subsection{Experiment 2 - Chronic Kidney Disease}

The term chronic kidney disease (CKD) describes all degrees of decreased renal function. It is more prevalent in the elderly population and it is estimated that affects 10–15\% of the world population. CKD is not often identified in premature stages.

The Chronic Kidney Dataset contains 25 features and a target that represents whether the patient has the Chronic Kidney Disease, for 400 patients, one third of which did not have the disease and two thirds that did. It is publicly available at the UC Irvine Machine Learning Repository \cite{dua.2019}.

The results of the experiments for this dataset are shown in Table \ref{results.kidney}. In this case, the aggregation using the inverse contribution does not improve the accuracy for all participants, since this metric worsens for the first uneven split. The size-based and contribution-base aggregation do not increase the accuracy of the local models for this split as well, while the Federated Averaging fails for the last uneven split and the size and accuracy weighted average for the second one. As in the previous experiment, the only aggregation method that impoves the accuracy for all cases is the accuracy-based one. 

\begin{table*}[ht]
\centering
\scriptsize
\caption{Experiment 2 - Chronic Kidney Disease}\label{results.kidney}
\tiny
\begin{tabular}{c | c | c|| c|| c|| c || c|| c|| c || c}
\hline
& & & \textbf{Acc} &         \textbf{Acc Federated} &        \textbf{Acc Size} &  \textbf{Acc Accuracy} & \textbf{Acc Size \& acc} &  \textbf{Acc Cont} & \textbf{Acc Inv Cont} \\
 \textbf{Part} & \textbf{Size} & \textbf{Pos (\%)}  & (\textbf{local}) & \textbf{Averaging} (\textbf{Fed}) & (\textbf{Fed}) &  (\textbf{Fed})  &  (\textbf{Fed}) & (\textbf{Fed}) &  (\textbf{Fed}) \\
\hline
1 & 20\% & 73\% & 1.0000 & 0.9375 &  1.0000 &  1.0000  & 1.0000  & 0.8750 &  0.8583\\
2 & 20\% & 58\% & 0.9375 & 1.0000 &  0.9375 &  1.0000  & 1.0000  & 1.0000 & 1.0000\\
3 & 20\% & 56\% & 0.8125 & 0.8750 & 0.9375 &  0.8750   & 0.9375  & 0.9375 &  0.8750\\
4 & 20\% & 65\% & 0.9375  & 1.0000 & 1.0000 &  0.8750  & 1.0000  & 1.0000 &  1.0000\\
5 & 20\% & 60\% & 1.0000 & 1.0000 & 0.8750 &  1.0000   & 0.8750  & 1.0000 &  1.0000 \\\hline
Avg & - & - &  0.9375 & \textbf{0.9625} & \textbf{0.9500} & \textbf{0.9500} & \textbf{0.9625} & \textbf{0.9625} & \textbf{0.9467} \\
\hline\hline
1 & 14\% & 36\% & 1.0000 & 1.0000 &  1.0000 & 0.9166 & 0.9000 & 0.9285 & 0.9166 \\
2 & 29\% & 67\% & 0.9583 & 0.9583 &  0.9583 & 1.0000 & 0.9545 & 0.8947 & 0.9565 \\
3 & 20\% & 77\% & 0.8750 & 0.9375 &  0.9285 & 1.0000 & 1.0000 & 0.9411 & 0.8571 \\
4 & 13\% & 48\% & 1.0000 & 1.0000 &  0.9000 & 1.0000 & 1.0000 & 1.0000 & 1.0000 \\
5 & 24\% & 68\% & 0.9500 & 0.9000 &  0.9583 & 0.9200 & 0.9545 & 1.0000 & 0.9523 \\\hline
Avg & - & - &  0.9567 & \textbf{0.9592} & 0.9490 & \textbf{0.9673} & \textbf{0.9618} & 0.9528 & 0.9365\\\hline\hline
1 & 52\% & 64\% & 1.0000 & 0.9761  & 0.9756 & 0.9250 & 0.9767  & 0.9750  & 0.9500 \\
2 &  3\% & 50\% & 1.0000 & 1.0000  & 1.0000 & 1.0000 & 0.5000  & 1.0000  & 1.0000 \\
3 & 14\% & 76\% & 1.0000 & 1.0000  & 1.0000 & 1.0000 & 0.9285  & 1.0000  & 0.9285 \\
4 &  5\% & 83\% & 0.7500 & 1.0000  & 1.0000 & 1.0000 & 1.0000  & 1.0000  & 1.0000 \\
5 & 26\% & 51\% & 0.9523 & 0.9523  & 0.9090 & 0.9523 & 0.8333  & 0.9500  & 0.9130 \\\hline
Avg & - & - & 0.9404 & \textbf{0.9857} & \textbf{0.9769} & \textbf{0.9755} & 0.8477 & \textbf{0.9855} & \textbf{0.9583} \\\hline\hline
1 & 44\% & 63\% &  0.9428 & 0.9428 & 0.9743 & 0.9750 &  0.9736 & 0.9047  & 0.9756\\
2 &  8\% & 72\% &  1.0000 & 1.0000 & 1.0000 & 1.0000 &  1.0000 & 1.0000  & 1.0000\\
3 &  6\% & 43\% &  0.8000 & 0.8000 & 1.0000 & 1.0000 &  1.0000 & 1.0000  & 1.0000\\
4 & 13\% & 43\% &  0.9090 & 0.9090 & 1.0000 & 1.0000 &  1.0000 & 1.0000  & 1.0000\\
5 & 30\% & 72\% &  1.0000 & 1.0000 & 0.8571 & 0.9444 &  1.0000 & 1.0000  & 1.0000\\\hline
Avg & - & - & 0.9304 & 0.9304 & \textbf{0.9663} & \textbf{0.9839} & \textbf{0.9947} & \textbf{0.9809} & \textbf{0.9951} \\\hline
\end{tabular}
\end{table*}

\subsection{Experiment 3 - Parkinson's}

Parkinson’s is a neurodegenerative disease that produces alterations in gait and posture that may increase the risk of falls and leads to mobility disabilities. Parkinson's affects about 1\% of the world population over the age of 55. 

The symptoms generally develop over years and their progressions are very diverse, making the diagnosis of the disease in the early stages extremely difficult. 

This dataset was created by Max Little of the University of Oxford \cite{Little07}, in collaboration with the National Centre for Voice and Speech, in Denver, Colorado and is composed by a range of biomedical voice measurements from patients. Each column in the table is a particular voice measure, and each row corresponds one of 195 voice recording from these individuals.

Table \ref{results.parkinson} shows the results of the different federation processes in this dataset. In this case, the splits are more polarized, finding that with the even split and the first uneven split all aggregation methods improve the performance of the local models, while for the second uneven split, with two participants with 4\% and 6\% of the data, no aggregation increases the accuracy. In the last uneven split, all aggregation methods improve the local models except the Federated Averaging. 

\begin{table*}[ht]
\centering
\scriptsize
\caption{Experiment 3 - Parkinson's}\label{results.parkinson}
\tiny
\begin{tabular}{c | c | c|| c|| c|| c || c|| c|| c || c}
\hline
& & & \textbf{Acc} &         \textbf{Acc Federated} &        \textbf{Acc Size} &  \textbf{Acc Accuracy} & \textbf{Acc Size \& acc} &  \textbf{Acc Cont} & \textbf{Acc Inv Cont} \\
 \textbf{Part} & \textbf{Size} & \textbf{Pos (\%)}  & (\textbf{local}) & \textbf{Averaging} (\textbf{Fed}) & (\textbf{Fed}) &  (\textbf{Fed})  &  (\textbf{Fed}) & (\textbf{Fed}) &  (\textbf{Fed}) \\
\hline
1 & 20\% & 65\% & 0.4285 & 0.7142 & 0.8571 & 1.0000 & 1.0000 & 1.0000 & 0.8571\\
2 & 20\% & 78\% & 0.7500 & 0.8750 & 1.0000 & 1.0000 & 0.8750 & 1.0000 & 1.0000\\
3 & 20\% & 75\% & 0.5000 & 0.8750 & 0.8750 & 0.8750 & 0.8750 & 0.8750 & 0.8750\\
4 & 20\% & 78\% & 0.7500 & 0.7500 & 0.8750 & 0.8750 & 0.8750 & 0.5000 & 0.7500\\
5 & 20\% & 79\% & 1.0000 & 1.0000 & 0.6250 & 0.8750 & 0.8750 & 0.8750 & 0.8750\\\hline
Avg & - & - & 0.6857 & \textbf{0.8429} & \textbf{0.8464} & \textbf{0.9250} &  \textbf{0.9000} & \textbf{0.8500} & \textbf{0.8714}\\
\hline\hline
1 & 11\% & 61\% & 0.6000 & 0.8000 &  0.6000  & 0.7142 & 0.8000 & 0.8000 & 0.6666\\
2 & 25\% & 71\% & 0.9000 & 0.9000 &  0.9000  & 0.7272 & 0.8461 & 0.8000 & 0.7777\\
3 & 27\% & 87\% & 0.8181 & 0.9090 &  1.0000  & 0.8750 & 1.0000 & 1.0000 & 0.8000\\
4 & 10\% & 45\% & 0.5000 & 0.7500 &  0.6000  & 0.7500 & 1.0000 & 0.7500 & 0.8000\\
5 & 27\% & 84\% & 0.9090 & 0.9090 &  0.8000  & 0.9090 & 0.8181 & 0.8333 & 1.0000\\\hline
Avg & - & - &  0.7454 & \textbf{0.8536}  & \textbf{0.7800} & \textbf{0.7951} & \textbf{0.8929} & \textbf{0.8367} & \textbf{0.8089}\\\hline\hline
1 & 50\% & 78\% & 0.8500 & 0.8500 & 0.8181  & 0.8500 & 0.7894  & 0.6666 & 0.8500 \\
2 &  6\% & 57\% & 0.6666 & 0.3333 & 1.0000  & 0.7500 & 0.6666  & 0.8000 & 0.7500 \\
3 & 13\% & 84\% & 0.7142 & 0.8571 & 0.7500  & 1.0000 & 0.3333  & 0.7142 & 0.8571 \\
4 &  4\% & 57\% & 1.0000 & 0.6666 & 0.6666  & 0.6666 & 0.7500  & 0.6666 & 0.3333\\
5 & 21\% & 79\% & 0.8750 & 0.8750 & 0.7777  & 0.6250 & 0.7777  & 0.6250 & 0.4285\\\hline
Avg & - & - &  0.8212 & 0.7164 & 0.8025 & 0.7783 & 0.6634  & 0.6945 & 0.6438\\\hline\hline
1 & 44\% & 63\% & 0.9428 & 0.9428 & 0.9743 &  0.9750 &  0.9736  & 0.9047  &  0.9756\\
2 &  8\% & 72\% & 1.0000 & 1.0000 & 1.0000 &  1.0000 &  1.0000  & 1.0000  &  1.0000\\
3 &  6\% & 43\% & 0.8000 & 0.8000 & 1.0000 &  1.0000 &  1.0000  & 1.0000  &  1.0000\\
4 & 13\% & 43\% & 0.9090 & 0.9090 & 1.0000 &  1.0000 &  1.0000  & 1.0000  &  1.0000\\
5 & 30\% & 72\% & 1.0000 & 1.0000 & 0.8571 &  0.9444 &  1.0000  & 1.0000  &  1.0000\\\hline
Avg & - & - & 0.9304 & 0.9304  & \textbf{0.9663} & \textbf{0.9839} & \textbf{0.9947} & \textbf{0.9810} & \textbf{0.9951}\\\hline
\end{tabular}
\end{table*}

\subsection{Experiment 4 - Heart Disease}

Heart disease describes a range of conditions that affect the heart, including blood vessel diseases, such as coronary artery disease, arrhythmia and congenital heart defects among others.

Heart disease is one of the biggest causes of morbidity and mortality among the population of the world. Prediction of cardiovascular disease is regarded as one of the most important subjects in the section of clinical data analysis. 

The Heart Disease dataset includes data from noninvasive test results of patients undergoing angiographies in order to study the possibility of angiographic coronary disease in them. The data collected include 14 attributes of the patients. For more information about this dataset, see \cite{Detrano89}.

In Table \ref{results.heart} we find the results of the different federation processes changing the aggregation method. With the even split we see that all aggregation methods improve the accuracy of the local models but for the aggregation based in a weighted average of the size of the participants and the accuracy of the local model. In the case of the first uneven split, the only aggregation methods that are able to improve the performance are the size-based and the accuracy-based. On the other hand, for the second uneven split, with two participants with 2\% and 4\% of the data, all aggregation methods increase the accuracy. Finally, for the last uneven split, only the Federated Averaging and the accuracy-based aggregation improve the performance of the models, resulting on the accuracy-based aggregation being the only aggregation that improves in all cases for this dataset.

\begin{table*}[ht]
\centering
\scriptsize
\caption{Experiment 4 - Heart Disease}\label{results.heart}
\tiny
\begin{tabular}{c | c | c|| c|| c|| c || c|| c|| c || c}
\hline
& & & \textbf{Acc} &         \textbf{Acc Federated} &        \textbf{Acc Size} &  \textbf{Acc Accuracy} & \textbf{Acc Size \& acc} &  \textbf{Acc Cont} & \textbf{Acc Inv Cont} \\
 \textbf{Part} & \textbf{Size} & \textbf{Pos (\%)}  & (\textbf{local}) & \textbf{Averaging} (\textbf{Fed}) & (\textbf{Fed}) &  (\textbf{Fed})  &  (\textbf{Fed}) & (\textbf{Fed}) &  (\textbf{Fed}) \\
\hline
1 & 20\% & 53\% & 0.7500 & 0.7500 & 1.0000 & 0.7500 &  0.8333 & 0.7500 & 0.9166   \\
2 & 20\% & 64\% & 0.6666 & 0.8333 & 0.8333 & 0.8333 &  0.7500 & 0.8333 & 1.0000    \\
3 & 20\% & 44\% & 0.5833 & 0.8333 & 1.0000 & 0.9166 &  0.5833 & 0.8333 & 0.8333  \\
4 & 20\% & 58\% & 0.9166 & 0.8333 & 0.7500 & 0.8333 &  0.5833 & 0.9166 & 0.9166  \\
5 & 20\% & 54\% & 0.9230 & 1.0000 & 0.7692 & 0.7692 &  0.6153 & 0.6923 & 0.6923  \\\hline
Avg & - & - & 0.7679 & \textbf{0.8500} & \textbf{0.8705} & \textbf{0.8205}  &  0.6731 & \textbf{0.8051} & \textbf{0.8718}\\
\hline\hline
1 & 15\% & 30\% & 0.8000 & 0.7000  & 1.0000  & 0.7777 & 0.9000 & 0.7142  & 0.6000\\
2 & 26\% & 63\% & 0.8750 & 0.7500  & 0.7500  & 0.7142 & 0.9375 & 0.9444  & 0.7647\\
3 & 18\% & 78\% & 0.9090 & 0.9090  & 1.0000  & 0.8333 & 0.7692 & 0.8888  & 0.7500\\
4 & 13\% & 39\% & 0.6250 & 0.8750  & 0.7777  & 0.8888 & 0.7000 & 1.0000  & 1.0000\\
5 & 28\% & 51\% & 0.8888 & 0.8333  & 0.9230  & 0.8888 & 0.5714 & 0.9047  & 0.8666\\\hline
Avg & - & - & 0.8195 & 0.8134  & \textbf{0.8902} & \textbf{0.8206} & 0.7756  & 0.7963 & 0.7963\\\hline\hline
1 & 50\% & 55\% & 0.6666 & 0.7333 & 0.7333  & 0.7500 &  0.7241 & 0.8387 & 0.7096   \\
2 &  2\% & 75\% & 0.5000 & 1.0000 & 0.6666  & 1.0000 &  0.5000 & 0.6666 & 0.6666 \\
3 & 14\% & 72\% & 0.8888 & 0.8888 & 0.8000  & 0.8750 &  0.8181 & 0.7272 & 0.7777  \\
4 &  4\% & 71\% & 0.3333 & 0.6666 & 0.3333  & 1.0000 &  0.6666 & 0.7500 & 0.5000  \\
5 & 30\% & 43\% & 0.6842 & 0.6315 & 0.8125  & 0.8235 &  0.5882 & 0.7142 & 0.7500   \\\hline
Avg & - & - & 0.6146 & \textbf{0.7841} & \textbf{0.6692} & \textbf{0.8897} & \textbf{0.6594} & \textbf{0.7394} & \textbf{0.6808}\\\hline\hline
1 & 47\% & 58\% & 0.7500 & 0.6875 & 0.8387 & 0.7500 & 0.7666 & 0.8928 & 0.8787 \\
2 &  7\% & 38\% & 0.8000 & 0.8000 & 0.6000 & 0.8000 & 0.6000 & 0.6666 & 0.6000 \\
3 & 10\% & 39\% & 0.4000 & 0.4000 & 0.2500 & 1.0000 & 0.2500 & 0.5000 & 0.7500 \\
4 &  9\% & 31\% & 0.8750 & 1.0000 & 1.0000 & 0.7142 & 0.6666 & 0.8000 & 0.6000 \\
5 & 26\% & 68\% & 0.9285 & 0.9285 & 0.8750 & 0.7857 & 0.9333 & 0.7777 & 0.8235 \\\hline
Avg & - & - & 0.7507 & \textbf{0.7632} & 0.7127 & \textbf{0.8100} & 0.6433 & 0.7274 & 0.7305
\\\hline
\end{tabular}
\end{table*}

\section{Conclusions}\label{conclusions}

Federated learning is a distributed artificial intelligence approach that can be very useful for hospitals to build collaboratively a machine learning model in a secure way, without sharing their private data. Nevertheless, the aggregation method is a decisive parameter that can change the performance of the final federated model. 

In this research we have proved that the classical Federated Averaging is a reliable aggregation method that improves the performance of the local methods in 11 out of 16 cases that we have contemplated. Nevertheless, there are other aggregation methods with similar or even better behaviour. The contribution-based aggregation, using the difference between the losses of the global and local model, increases the accuracy in 11 out of 16 cases as well, while the size-based and the inverse contribution-based perform better in one more case. The weighted average using both the size of the participant's dataset and the accuracy of the local model increase the accuracy in only 10 cases out of 16. Finally, the weighted average using the accuracy outperforms all aggregation methods, improving the accuracy in 15 out of 16 cases, that is, in all experiments but one partition where all other methods failed to increase the performance as well. 

With this results, the authors believe that an accuracy-based federated learning may perform better than the Federated Averaging classical approach. A fully connected peer-to-peer architecture has been used to show a resilient architecture against different points of failures, including the central server. As limitations of this study, using open-source datasets instead of real-world data in a study on federated learning with medical data may limit the realism, generalization, diversity, quality, and ethical considerations of the research findings.

\section*{Acknowledgements} Prof. Salmeron research was kindly supported by the project Artificial Intelligence for Healthy Aging (Convocatoria 2021 – Misiones de I+D en Inteligencia Artificial: Inteligencia Artificial distribuida para el diagnóstico y tratamiento temprano de enfermedades con gran prevalencia en el envejecimiento, exp.: MIA.2021.M02.0007) lead by Capgemini Engineering.

Author contribution statement:

Jose L. Salmeron; Irina Arevalo; Antonio Ruiz-Celma: Conceived and designed the experiments; Performed the experiments; Analyzed and interpreted the data; Contributed reagents, materials, analysis tools or data; Wrote the paper. 

Data availability statement: 

Data included in article/supp. material/referenced in article.

\bibliographystyle{plain}

\bibliography{main}

\end{document}